\documentclass[runningheads]{llncs}
\usepackage{graphicx}
\usepackage{todonotes}
\usepackage{multirow}
\usepackage{booktabs}

\usepackage{array}
\usepackage{tabu}
\newcolumntype{M}[1]{>{\centering\arraybackslash}m{#1}}
\newcolumntype{L}[1]{>{\arraybackslash}m{#1}}

\begin{document}
\title{Overview of BioASQ 2025: The thirteenth BioASQ challenge on large-scale biomedical semantic indexing and question answering
}
    \titlerunning{Overview of BioASQ 2025}

\author{
Anastasios Nentidis\inst{1} \and
Georgios Katsimpras\inst{1} \and
Anastasia Krithara\inst{1} \and
Martin Krallinger\inst{2} \and
Miguel Rodríguez-Ortega\inst{2} \and
Eduard Rodriguez-López\inst{2} \and
Natalia Loukachevitch\inst{3} \and
Andrey Sakhovskiy\inst{5,6} \and
Elena Tutubalina\inst{4,5}\and
Dimitris Dimitriadis\inst{7}\and
Grigorios Tsoumakas\inst{7,10}\and
George Giannakoulas\inst{7}\and
Alexandra Bekiaridou\inst{8}\and
Athanasios Samaras\inst{7}\and
Giorgio Maria Di Nunzio\inst{9}\and 
Nicola Ferro\inst{9}\and 
Stefano Marchesin\inst{9}\and 
Marco Martinelli\inst{9}\and 
Gianmaria Silvello\inst{9}\and
Georgios Paliouras\inst{1}
}
\authorrunning{A. Nentidis et al.}
%
\institute{
National Center for Scientific Research ``Demokritos'', Athens, Greece\\
\email{\{tasosnent, gkatsibras, akrithara,  paliourg\}@iit.demokritos.gr} \and
Barcelona Supercomputing Center, Barcelona, Spain\\
\email{\{martin.krallinger,  mirodrig8,  eduard.rodriguez\}@bsc.es}
\and
Moscow State University, Russia \\
\email{louk\_nat@mail.ru}
\and
Artificial Intelligence Research Institute, Russia
\and
Kazan Federal University, Russia\\
\and
SberAI \& Skoltech, Russia\\
\email{\{andrey.sakhovskiy, tutubalinaev\}@gmail.com} \and
Aristotle University of Thessaloniki, Greece\\
\email{\{dndimitri,greg\}@csd.auth.gr, \{g.giannakoulas, th.samaras.as\}@gmail.com}
\and
Northwell Health, USA\\
\email{ampekiaridou@gmail.com}
\and
University of Padua, Italy\\
\email{\{name.surname\}@unipd.it} 
\and
Archimedes, Athena Research Center, Greece
}
\maketitle              
\begin{abstract}
This is an overview of the thirteenth edition of the BioASQ challenge in the context of the Conference and Labs of the Evaluation Forum (CLEF) 2025. 
BioASQ is a series of international challenges promoting advances in large-scale biomedical semantic indexing and question answering. 
This year, BioASQ consisted of new editions of the two established tasks, b and Synergy, and four new tasks: 
a) \textit{Task MultiClinSum} on multilingual clinical summarization. 
b) \textit{Task BioNNE-L} on nested named entity linking in Russian and
English. 
c) \textit{Task ELCardioCC} on clinical coding in cardiology.
d) \textit{Task GutBrainIE} on gut-brain interplay information extraction.
In this edition of BioASQ, 83 competing teams participated with more than 1000 distinct submissions in total for the six different shared tasks of the challenge. 
Similar to previous editions, several participating systems achieved competitive performance, indicating the continuous advancement of the state-of-the-art in the field.

\end{abstract}
\section{Introduction}
The BioASQ challenge was introduced over a decade ago, aiming to advance the state-of-the-art in large-scale biomedical semantic indexing and question answering (QA)~\cite{Tsatsaronis2015}.
To achieve this, it hosts annual shared tasks, creating benchmark datasets that reflect the real-world information needs of biomedical experts. These include new versions of established tasks that remain relevant and timely, as well as novel tasks introduced to explore and address unmet biomedical information needs.     
These shared tasks provide research teams worldwide, who are developing systems for biomedical semantic indexing and QA, with access to publicly available datasets, a standardized evaluation framework, and opportunities for knowledge exchange through the BioASQ challenge and workshop.

Here, we present the shared tasks and the datasets of the thirteenth edition of the BioASQ challenge in 2025, as well as a condensed overview of the participating systems and their performance.
The remainder of this paper is organized as follows. 
First, Section~\ref{sec:tasks} presents a general description of the shared tasks, which took place in 2025, and the corresponding datasets developed for the challenge. 
Then, Section~\ref{sec:participants} provides a brief overview of the participating systems for the different tasks. 
Detailed descriptions for some of the systems are available in the respective extended overviews of each task and the proceedings of the BioASQ lab. 
Subsequently, in Section~\ref{sec:results}, we present the performance of the systems for each task, based on state-of-the-art evaluation measures or manual assessment.
Finally, in Section~\ref{sec:conclusion} we draw some conclusions.

\section{Overview of the tasks}
\label{sec:tasks}

The thirteenth edition of the BioASQ challenge consisted of six tasks~\cite{BioASQECIR2025}:
(i) \textit{Task b} on biomedical semantic question answering. 
(ii) \textit{Task Synergy} on question answering developing biomedical topics.
(iii) \textit{Task MultiClinSum} on multilingual clinical summarization. 
(iv) \textit{Task BioNNE-L} on nested named entity linking in Russian and
English. 
(v) \textit{Task ELCardioCC} on clinical coding in cardiology.
(vi) \textit{Task GutBrainIE} on gut-brain interplay information extraction.
In this section, we first describe this year's editions of the two established tasks b (task 13b) and Synergy (Synergy 13)~\cite{BioASQ2024task12bSynergy} with a focus on differences from previous editions of the challenge~\cite{nentidis2023results,nentidis2022overview}. Additionally, we also introduce the four new BioASQ tasks, MultiClinSum~\cite{BioASQ2025MultiClinSum}, BioNNE-L~\cite{bionnel-overview-2025}, ELCardioCC~\cite{BioASQ2025ElCardioCC}, and GutBrainIE~\cite{BioASQ2025taskGutBrainIE}.

\subsection{Task 13b}

BioASQ \textit{task 13b} is the thirteenth edition of the established BioASQ \textit{task b} on Biomedical QA~\cite{BioASQ2025task13bSynergy}. This year, it took place in three phases: i) Phase A: biomedical questions in English were provided, and the systems had to retrieve relevant material (PubMed documents and snippets). ii) Phase A+, the systems had to provide ‘exact’ and ‘ideal’ answers. Depending on question type, the ‘exact’ answer can be a \textit{yes} or \textit{no} (yes/no), an entity name, such as a disease or gene (factoid), or a list of entity names (list). The ‘ideal’ answer is a paragraph-sized summary, regardless of question type. iii) Phase B: Some relevant material was provided for each question, selected by the BioASQ experts, and the systems had to provide new answers given this additional information. 

About 340 new biomedical questions annotated with golden documents, snippets, and answers (‘exact’ and ‘ideal’), were developed for testing.
In addition, a training set of 5,389 biomedical questions, accompanied by answers, and supporting evidence (documents and snippets), was available from previous versions of the tasks, as a unique resource for the development of question-answering systems \cite{krithara2023bioasq}.
Table \ref{tab:b_data} presents some statistics of both training and test datasets for task 13b.
The test data for task 13b were split into four independent bi-weekly batches consisting of 85 questions each, as presented in Table \ref{tab:b_data}. 

\begin{table}[!htb]
        \caption{Statistics on the training and test datasets of task 13b. The numbers for the documents and snippets refer to averages per question.}\label{tab:b_data}
        \centering
        \begin{tabular}{M{0.09\linewidth}M{0.08\linewidth}M{0.08\linewidth}M{0.09\linewidth}M{0.15\linewidth}M{0.15\linewidth}M{0.15\linewidth}M{0.15\linewidth}}\hline
        \textbf{Batch} 	& \textbf{Size} 	&	\textbf{Yes/No}	&\textbf{List}	&\textbf{Factoid}	&\textbf{Summary}& \textbf{Documents} 	& \textbf{Snippets}  	\\\hline
        Train  & 5389 & 1459  & 1047 & 1600    & 1283    & 9.74   & 12.78      \\
        Test 1         & 85   & 17    & 23   & 26      & 19      & 2.68   & 3.74       \\
        Test 2         & 85   & 17    & 19   & 27      & 22      & 2.71   & 3.06       \\
        Test 3         & 85   & 22    & 22   & 20      & 21      & 3.00   & 3.66       \\
        Test 4         & 85   & 26    & 19   & 22      & 18      & 3.15   & 3.92       \\\hline
        \textbf{Total} & 5729 & 1541  & 1130 & 1695    & 1363    & 9.33   & 12.23    \\\hline
        \end{tabular}
\end{table}

\subsection{Task Synergy 13}

BioASQ \textit{task Synergy} was originally introduced in 2020 with the aim of promoting research in developing biomedical topics, such as COVID-19~\cite{krithara2021bioasq,Synergy_jamia}.
The design of this task as an ongoing dialogue allows experts to pose open-ended questions for developing topics, for which they do not know in advance whether a definitive answer can be given, in order to obtain relevant material (documents and excerpts) retrieved from the systems. After assessing this material, they provide feedback to the systems on its relevance and on whether it is sufficient to answer their question, by marking respective questions as \textit{ready to answer}. This process is repeated iteratively in rounds with new material considered in each round, based on updates to the original document resource~\cite{nentidis2021overview}\footnote{As of 2023, this evolving document resource is PubMed~\cite{nentidis2023results}}.
For \textit{ready to answer} questions, they receive exact and ideal answers as well, assess them, and provide feedback that can be used by the systems to improve their responses to these questions in the remaining rounds. 
The experts can also mark a question as \textit{closed} if they receive a fully satisfactory answer that is not expected to change or if they are no longer interested in the question.

A training dataset of 366 questions on developing topics with incremental annotations with relevant material and answers is already available from previous versions of \textit{task Synergy}~\cite{BioASQ2024task12bSynergy,nentidis2023ceur,nentidis2022ceur,nentidis2021ceur}. 
During the \textit{task Synergy 13}, this set was extended with 47 new questions on developing health topics, such as infectious, rare, and genetic diseases, and women’s and reproductive health. Meanwhile, 27 questions from the previous version of the task remained open and were enriched with more recent evidence and updated answers~\cite{BioASQ2025task13bSynergy}. 
Overall, 74 questions were considered in the four rounds of \textit{task Synergy 13}. The number of yesno, list, factoid, and summary questions was 23, 19, 14, and 18, respectively.
 
\subsection{Task MultiClinSum}

There is a rapid accumulation of various types of clinical content, including medical records and publications such as clinical case reports, written not only in English but also in many other languages. Some clinical reports can be very lengthy, making it challenging for healthcare professionals and even patients to comprehend and extract key clinical insights. Large Language Models (LLMs) have shown promising results in automatic summarization, helping to condense lengthy clinical documents into shorter versions or summaries that retain the most relevant clinical information. Therefore, there is a pressing need to evaluate and benchmark the performance of different clinical summarization methods, especially for content written in multiple languages.

We introduce the \textit{MultiClinSum} task covering the automatic summarization of lengthy clinical case reports written in English, Spanish, French, and Portuguese. The \textit{MultiClinSum} task relies on a corpus of manually selected full clinical case reports with their corresponding summaries derived from case report publications written in the mentioned languages. This gold standard dataset comprises 1,280 pairs of full-text and summary in English, 534 in Spanish, 200 in Portuguese and 200 in French. To increase the size of the dataset, both full-text and summary texts of each language were translated with neural machine translation models into the other languages, resulting in a total of 1,976 pairs for each language. An additional large-scale dataset was also created derived from the PMC-Patients clinical cases (full-text) and their corresponding summary extracted from the PubMed abstracts. Table ~\ref{tab:multiclinsum_corpus-overview} shows the corpus statistics for each sub-track of MultiClinSum.

For the evaluation assessment, the automatically generated summaries were compared with the summaries that had been manually generated by the original authors, using Rouge-L \cite{Lin2004} scores and BERTScore \cite{zhang2020bertscore}. As clinical case reports do share commonalities with medical discharge summaries (patient demographics, relevant medical history, clinical presentation, diagnostic process, intervention, treatment, outcome and follow-up), insights provided by the \textit{MultiClinSum} results can be of practical relevance also for clinical records summarization scenarios.

\begingroup
\setlength{\tabcolsep}{4pt}
\begin{table}[]
    \centering
    \caption{Statistics for the datasets provided for MultiClinSum indicating the number of fulltext-summary pairs available for each sub-track.}
    \begin{tabular}{lccccccM{0.12\linewidth}}
        \hline
        \textbf{Sub-track} & \textbf{Lang.} & \textbf{Dataset} & \textbf{Fulltext-summary pairs}  \\
        \hline
        MultiClinSum-gs-en & EN & Native/Transl. gold stand. & 988   \\
        MultiClinSum-gs-es & ES & Native/Transl. gold stand. & 988   \\
        MultiClinSum-gs-fr & FR & Native/Transl. gold stand. & 1061    \\
        MultiClinSum-gs-pt & PT & Native/Transl. gold stand. & 1034   \\
        \hline
        MultiClinSum-ls-es & EN & Large Scale (PMC-Patients)& 28.902   \\
        MultiClinSum-ls-es & ES & Large Scale (PMC-Patients)& 28.902    \\
        MultiClinSum-ls-fr & FR & Large Scale (PMC-Patients)& 28.902    \\
        MultiClinSum-ls-pt & PT & Large Scale (PMC-Patients)& 28.902   \\
        \hline
    \end{tabular}
    \label{tab:multiclinsum_corpus-overview}
\end{table}
\endgroup

\subsection{Task BioNNE-L}
In the BioNNE-L Shared Task~\cite{bionnel-overview-2025}, we address the medical entity linking task, also known as Medical Concept Normalization (MCN), which is to map given entities to the most relevant vocabular entries from an external source, e.g., concepts from the UMLS metathesaurus~\cite{bodenreider2004unified} identified with concept unique identifiers (CUIs). Although the task has been widely explored in recent years, existing approaches usually treat each entity individually, medical entities often form a nested structure, where an entity can be a subpart of another entity. One of the key features of BioNNE-L is the focus on nested entities that are (i) derived from the MCN annotation of the NEREL-BIO corpus~\cite{NERELBIO,NEREL-BIO-COLING-2024} and (ii) supplemented by newly annotated data in both English and Russian. The annotated entity types  are disorders (\textit{DISO}), anatomical structures (\textit{ANAT}), and chemicals (\textit{CHEM}) normalized to UMLS. The competition was organized into three subtasks that fell under two evaluation tracks: 1. \textbf{Monolingual track} that treated English and Russian data independently; 2. \textbf{Bilingual track} that required a single bilingual model for the combined Russian and English data. Data statistics for both tracks, as well as the normalization dictionary, are summarized in Table~\ref{tab:bionnel-data-statistics}.

All BioNNE-L materials can be found on the shared task's GitHub\footnote{\url{https://github.com/nerel-ds/NEREL-BIO/tree/master/BioNNE-L\_Shared\_Task}} and Codalab pages\footnote{\url{https://codalab.lisn.upsaclay.fr/competitions/21568}}. Annotated data and normalization dictionary are also available at HuggingFace\footnote{\url{https://huggingface.co/datasets/andorei/BioNNE-L}}.

\begin{table}[!htb]
\setlength{\tabcolsep}{3.5pt}
\centering
 \caption{BioNNE-L 2025 statistics for Disorder (\textbf{DISO}), Chemical (\textbf{CHEM}), and Anatomical Structure (\textbf{ANAT}) among Russian and English entities as well as normalization dictionary statistics.}
\begin{tabular}{lc|c|c|c|c|c|c|c}\hline

\textbf{Entity} & \multicolumn{4}{c|}{\textbf{Refined NEREL-BIO}} & \multicolumn{2}{c|}{\textbf{Novel data}} & \multicolumn{2}{c}{\textbf{Dictionary}} \\
\cmidrule{2-5}\cmidrule{6-7}\cmidrule{8-9}
\textbf{type} & \multicolumn{2}{c|}{\textbf{Train}} & \multicolumn{2}{c|}{\textbf{Dev}} & \multicolumn{2}{c|}{\textbf{Test}} & & \\
& \textbf{Ru}  & \textbf{En} & \textbf{Ru}  & \textbf{En} & \textbf{Ru}  & \textbf{En} & \textbf{Ru}  & \textbf{En} \\
\midrule
\# documents & 716 & 54 & 50 & 50 & 154 & 154 & --- & --- \\
\midrule
\multicolumn{9}{c}{\textbf{Number of entities}} \\
\midrule
DISO & 11,168 & 1,200 & 925 & 1,029 & 2,811 & 3,068 & 91,867 & 1,825,048 \\
CHEM & 4,741 & 579 & 531 & 564 & 1,218 & 1,345 & 47037 & 1,732,096 \\
ANAT & 8,346 & 911 & 878 & 901 & 2,186 & 2,248 & 6899 & 345,043 \\
\hline
& 24,255 & 2,690 &  2,334 & 2,494 & 6,215 & 6,661 & 145,803 & 3,902,187 \\
\hline
\end{tabular}
\label{tab:bionnel-data-statistics}
\end{table}

\subsection{Task ELCardioCC}

Cardiovascular diseases affect a significant portion of the global population, accounting for 32\% of global deaths according to WHO\footnote{\url{https://www.who.int/health-topics/cardiovascular-diseases}}. Automated clinical coding plays a crucial role in transforming unstructured real-world medical data gathered from patients into structured information, in order to facilitate clinical research and analysis. However, existing research predominantly focuses on English clinical text, leaving other languages, such as Greek, underrepresented. To this end, we propose a new \textit{ELCardioCC} task \cite{BioASQ2025ElCardioCC}, which concerns i) the assignment of cardiology-related ICD-10 codes to discharge letters from Greek hospitals, ii) the extraction of the specific mentions of ICD-10 codes from the discharge letters.  

In detail, the participants in the \textit{ELCardioCC} task  were tasked with developing named entity recognition (NER), entity linking (EL) and multi-label classification - explainable AI (MLC-X) systems using a specialized corpus of discharge letters. These discharge letters, which were written in Greek contained valuable medical information about patients’ conditions, treatments, and outcomes. The corpus was meticulously annotated with the positions of mentions (such as chief complaint, diagnosis, prior medical history, drugs and cardiac echo) and their corresponding ICD-10 codes. The training dataset includes 1,000 discharge letters, while the test set comprises 500 letters. System performance was evaluated using the micro F1 score.

\subsection{Task GutBrainIE}
Recent scientific evidence suggests a connection between \emph{brain-related diseases} and the \emph{gut microbiota} that may play a critical role in mental health-related disorders or diseases like Parkinson’s, and Alzheimer’s \cite{appleton2018gut,carabotti2015gut,cryan2020gut,ghaisas2016gut}. The scientific literature on this topic is rapidly expanding, making it increasingly challenging for clinicians and researchers to stay up to date. For example, in 2020, approximately 200 articles were published on the relationship between gut microbiota and mental health; by 2024, this number had more than doubled to over 450 publications.
The \textit{GutBrainIE} Task aims to foster the development of Information Extraction (IE) systems that support experts by automatically extracting and linking knowledge from biomedical abstracts, facilitating the understanding of gut-brain interplay and its role in mental health and neurological diseases. 

The \textit{GutBrainIE} Task comprises four subtasks of increasing difficulty: Named Entity Recognition (NER), which identifies and classifies entity mentions in PubMed abstracts about the gut-brain interplay focusing on mental health and the Parkinson's disease; Binary Tag-based Relation Extraction (BT-RE), which detects whether pairs of entities are in relation without specifying the relation type; Ternary Tag-based Relation Extraction (TT-RE), which extends BT-RE by also assigning a relation label to each related pair; and Ternary Mention-based Relation Extraction (TM-RE), which further localizes the exact entity mentions involved in each relation and assigns the appropriate relation label.

The dataset includes over 1000 documents with annotated entity mentions and relations, organized into Training, Development, and Test sets. 
The train set is further divided into four quality tiers: expert-curated (Platinum), expert-annotated (Gold), student-annotated (Silver), and automatically generated (Bronze). 
Development and Test sets contain only expert annotations (Platinum+Gold).

\begin{table}[!htb]
    \centering
    \caption{Dataset statistics for \textit{GutBrainIE}. 
    }
    \label{tab:gutbrainie_dataset_summary}
    \begin{tabular}{|l|r|r|r|r|r|}
    \hline
    \multicolumn{1}{|l|}{\textbf{Collection}} & 
    \multicolumn{1}{c|}{\textbf{\# Docs}} & 
    \multicolumn{1}{c|}{\textbf{\# Entities}} & 
    \multicolumn{1}{c|}{\textbf{Ents/Doc}} & 
    \multicolumn{1}{c|}{\textbf{\# Rels}} & 
    \multicolumn{1}{c|}{\textbf{Rels/Doc}} \\
    \hline
    Train Platinum  & 111   & 3638  & 32.77 & 1455  & 13.11 \\
    Train Gold      & 208   & 5192  & 24.96 & 1994  & 9.59  \\
    Train Silver    & 499   & 15275 & 30.61 & 10616 & 21.27 \\
    Train Bronze    & 749   & 21357 & 28.51 & 8165  & 11.90 \\\hline
    Development Set & 40    & 1117  & 27.93 & 623   & 15.58 \\\hline
    Test Set        & 40    & 1237  & 30.92 & 777   & 19.42 \\
    \hline
    \end{tabular}
\end{table}

\section{Overview of participation}
\label{sec:participants}

Overall, 83 distinct teams participated in the thirteenth edition of the BioASQ challenge, submitting more than 1000 distinct runs for the six different shared tasks of the challenge. The majority of the teams focused on a single task, still some of them participated in two or even three BioASQ tasks\footnote{In particular, two teams participated in 13b \& Synergy13, one in 13b \& MultiClinSum, one in 13b \& GutBrainIE, one in 13b, MultiClinSum, \& ElCardioCC, and one in BioNNE-L, ElCardioCC, \& GutBrainIE}.  
In this section, we provide a condensed overview of the methods developed by the participating teams for each of the BioASQ tasks. However, a more detailed overview of these methods will be available in the extended overview of each task~\cite{BioASQ2025task13bSynergy,BioASQ2025MultiClinSum,bionnel-overview-2025,BioASQ2025ElCardioCC,BioASQ2025taskGutBrainIE}, and some method-specific descriptions will be available in the
proceedings of the thirteenth BioASQ workshop\footnote{\url{https://www.bioasq.org/workshop2025/proceedings}}.

\subsection{Task 13b}

This year, 46 teams participated in task 13b, submitting a total of 734 different submissions generated by 146 distinct systems across all four batches for the three phases A, A+, and B. 
This corresponds to a significant increase in participation, compared to the 26 teams in the previous version of the task (12b)\cite{BioASQ2024task12bSynergy}, which highlights that the task remains timely and relevant.   
Specifically, 34, 20, and 26 teams competed in phases A, A+, and B of task 13b, with 95, 79, and 88 distinct systems, respectively. Eleven of these teams were involved in all three phases.
As in previous years, the open-source system OAQA~\cite{yang2016learning}, which achieved top performance in older editions of BioASQ~\cite{Krithara2016overview}, was used as a baseline for phase B \textit{exact answers}.

The participating teams employed a range of well-established and sophisticated techniques. Many teams utilized traditional document retrieval methods such as BM25 and dense retrieval models (e.g. BGE-M3 and MiniLM), often improving results with re-ranking techniques. Some teams incorporated Retrieval-Augmented Generation (RAG) frameworks, using Large Language Models (LLMs) such as Llama, Gemma, GPT, Claude, and Mistral to generate responses. Beyond these methods, the teams also experimented with self-feedback mechanisms, zero-shot and few-shot prompting, ensemble methods, and the integration of biomedical knowledge bases to improve overall performance~\cite{panou_2025,Verma25,Stachura25,Chen25,Kim25,Galat25,Tang25,Angulo25,Ateia25,Jonker25,Borazio25}.

\subsection{Task Synergy 13}

In the thirteenth edition of BioASQ, five teams participated in the Synergy task (Synergy 13). These teams submitted 46 runs from 21 distinct systems. 
Two of these teams participated in task 13b as well, while the remaining three focused exclusively on task Synergy 13. 
The participating teams primarily utilized LLMs, such as DeepSeek-R1 and Llama. To further enhance performance, the teams experimented with RAG frameworks and employed techniques such as optimized prompting, NER, and majority voting to refine their results~\cite{Duenas_Romero,panou_2025}.
More detailed descriptions for some of the systems are available at the proceedings of the workshop.

\subsection{Task MultiClinSum}

In general, there has been a very satisfactory participation in the task with promising results in each of the presented sub-tracks. 56 teams registered for the MultiClinSUM task, out of which 11 teams submitted at least one run of their predictions. Specifically, 7 teams participated in the English sub-track, 5 teams in the Spanish, 4 teams in French, and 5 in Portuguese. Each team was allowed to submit up to 5 runs per sub-track. As expected, the best results were obtained in the English sub-track (MultiClinSum-en), which had the highest level of participation. Nevertheless, the others sub-tracks were quite well represented in terms of both participation and novel methodologies applied~\cite{BioASQ2025MultiClinSum}.

\subsection{Task BioNNE-L}

In total, we've received 23 Codalab registrations  for the BioNNE-L task, with 7 teams submitting predictions during the evaluation phase. The systems submitted by the participants are summarized in Table~\ref{tab:bionnel-team-overview}.

\begin{table}[!htb]
\centering
 \caption{Overview of the approaches presented by participants for the BioNNE task. EN stands for the English-oriented and RU for the Russian-oriented tracks.}
\begin{tabular}{lcp{5.5cm}}\hline
\textbf{Team} & \textbf{Track} & \textbf{Approach} \\ \hline
 verbanexialab  & EN & SapBERT w/ lexical and semantic reranking \\
 LYX\_DMIIP\_FDU   & Bilingual,EN,RU & BERGAMOT fine-tuning \\
 BlancaPlanca   & Bilingual,EN,RU & BERGAMOT w/ language-specific preprocessing \\
 MSM Lab   & Bilingual,EN,RU & Two-step retrieval and ranking pipeline \\
 dstepakov   & Bilingual,RU & RoBERTa fine-tuning with contrastive learning \\
 ICUE   & Bilingual,EN,RU & BERT, BioSyn, LLM 0-shot reranking \\
 NLPIMP   & Bilingual & Russian LaBSE model pre-trained on medical data \\

 \hline
 \end{tabular}
\label{tab:bionnel-team-overview}
\end{table}
Team \textbf{verbanexialab}~\cite{bionnel-VerbaNex-AI-Lab} leveraged a SapBERT\footnote{\url{https://huggingface.co/cambridgeltl/SapBERT-from-PubMedBERT-fulltext}}~\cite{sapbert}, pre-trained on UMLS concepts, to obtain entity embeddings, followed by a multicomponent re-ranking. They combined embedding cosine similarity with Jaccard similarity for lexical overlap recognition and Levenshtein distance for character-level alignment.

Team \textbf{LYX\_DMIIP\_FDU}~\cite{bionnel-LYX-DMIIP-FDU} fine-tuned a BERGAMOT\footnote{\url{https://huggingface.co/andorei/BERGAMOT-multilingual-GAT}}~\cite{BERGAMOT} model for each task via contrastive learning using the train- and dev-set entities to enrich the original vocabularies. The textual context of each entity was used as additional input to enhance the entity representation.

Team \textbf{BlancaPlanca}~\cite{bionnel-burlova} used BERGAMOT for zero-shot retrieval based on entity-concept cosine similarity. They apply language-specific lemmatization for Russian and speed up the inference by chucking the normalization dictionary into type-specific parts of 100k entries each.

Team \textbf{MSM Lab}~\cite{bionnel-MSM-Lab} adopted SapBERT~\cite{sapbert,sapbert-multilingual} and BioMedBERT~\cite{pubmedbert} for two-step retrieval and ranking.

Team \textbf{dstepakov} performed the nearest-neighbor search based on the cosine similarity of RoBERTa embeddings~\cite{roberta}, fine-tuned contrastively on anchor-positive-negative term triplets via the InfoNCE objective~\cite{Oord2018RepresentationLW}.

Team \textbf{ICUE}~\cite{bionnel-ICUE} fine-tuned BioSyn~\cite{biosyn} using the vocabularies reduced to less than 100k entries each. They fine-tune a separate BERT-based model~\cite{Devlin2018} for English~\cite{scibert}, Russian\footnote{\url{https://huggingface.co/KoichiYasuoka/bert-base-russian-upos}}, and multilingual~\cite{tedeschi-etal-2021-wikineural-multilingual-ner} tracks, respectively. They re-ranked the initial retrieval results using \textit{DeepSeek-R1-Distill-Llama-8B}\footnote{\url{https://huggingface.co/deepseek-ai/DeepSeek-R1-Distill-Llama-8B}}.

Team \textbf{NLPIMP}  performed the zero-shot ranking using a Russian LaBSE~\cite{feng-etal-2022-language} model\footnote{\url{https://huggingface.co/sergeyzh/LaBSE-ru-turbo}} pre-trained contrastively on an in-house Russian medical corpus.

\subsection{Task ELCardioCC}

The ELCardioCC task engaged five teams across its subtasks: NER, EL, and MLC-X, with a total of 13-14 systems submitted for each subtask in addition to baseline models. Most participating systems predominantly utilized transformer-based architectures, especially BERT variants and large multilingual language models (LLMs), for all three tasks. Common approaches included fine-tuning models like Greek BERT and XLM-Roberta for NER, employing semantic similarity with embedding models for EL, and using LLMs for classification and justification in MLC-X, often leveraging cross-lingual techniques to process Greek medical texts.

The ELCardioCC baselines, designed for clarity and reproducibility, primarily used multilingual BERT models adapted for each specific task. The NER baseline involved a fine-tuned cased mBERT model with BIO2 tagging. For EL, a context-aware hierarchical classifier built on mBERT was used, reflecting the ICD-10 taxonomy. The MLC-X baseline employed a Greek-BERT model for multi-label classification of the 40 most frequent ICD-10 codes, with variations for document-level predictions and rule-based justification of code selections.


\subsection{Task GutBrainIE}
The  \textit{GutBrainIE} task registered 17 teams submitting runs. Among these, 16 teams participated in NER, 12 in BT-RE, 13 in TT-RE, and 13 in TM-RE. 
Overall, a total of 391 runs were submitted: 101 for NER, 100 for BT-RE, and 95 for both TT-RE and TM-RE. 

Most teams adopted supervised fine-tuning or transformer-based models pre-trained on biomedical text for the NER task \cite{team_Gut-Instincts_paper_gbie,team_graphwise-1_paper_gbie,team_GutUZH_paper_gbie,team_BIU-ONLP_paper_gbie,team_ICUE_paper_gbie,bionnel-LYX-DMIIP-FDU,team_DS-GT-bioasq-task6_paper_gbie,team_ataupd2425-pam_paper_gbie,team_ataupd2425-gainer_paper_gbie,team_NLPatVCU_paper_gbie}. 
Standard backbones included PubMedBERT, BioBERT, BioLinkBERT, and ELECTRA \cite{clark2020electra,pubmedbert,lee2019biobert,linkbert}. 
Specialized NER architectures, such as GLiNER \cite{zaratiana-etal-2024-gliner}, were also utilized and fine-tuned. 
Many groups trained multiple models with different random seeds to boost robustness and ensembled their outputs. 
All teams used platinum, gold, and silver collections for training. 
A few also used the noisier bronze set, employing cleaning or re-weighting approaches and integrating PubMed data augmentation.

Across the RE subtasks, participants primarily used biomedical pre-trained language models fine-tuned on entity-marker augmented inputs \cite{team_Gut-Instincts_paper_gbie,team_graphwise-1_paper_gbie,team_GutUZH_paper_gbie,team_BIU-ONLP_paper_gbie,team_ICUE_paper_gbie,bionnel-LYX-DMIIP-FDU,team_DS-GT-bioasq-task6_paper_gbie,team_ataupd2425-pam_paper_gbie,team_ataupd2425-gainer_paper_gbie,team_NLPatVCU_paper_gbie}. 
Among these, the most widely employed include: SapBERT, PubMedBERT, BioBERT, RoBERTa, and ELECTRA \cite{sapbert,pubmedbert,lee2019biobert,roberta,clark2020electra}. 
Several teams reformulated RE as a seq2seq problem using REBEL-large \cite{huguet-cabot-navigli-2021-rebel-relation}, directly generating relation tuples or tagged spans in a single pass for all three subtasks. 
Few others leveraged model ensembling and trained with negative-pair subsampling to counter class imbalance and increase generalization capabilities. 
Finally, some groups experimented with few-shot or Retrieval-Augmented Generation (RAG), prompting large language models to extract both entities and relations with minimal fine-tuning \cite{team_lasigeBioTM_paper_gbie,team_greenday_paper_gbie,team_ONTUG_paper_gbie,team_ICUE_paper_gbie}. 

\section{Results}
\label{sec:results}
\subsection{Task 13b}

This section presents the evaluation measures and preliminary results for Task 13b.  
The evaluation in \textit{task 13b} is done manually by the experts that assess each system response and automatically by employing a variety of established evaluation measures \cite{malakasiotis2020evaluation} as in the previous versions of the task~\cite{BioASQ2024overview}. Table~\ref{tab:b_eval} provides a brief overview of the official measures per response and question type. 
The results reported for \textit{task 13b} are preliminary, as the final results will be available after the manual assessment of all system responses by the BioASQ team of experts and the enrichment of the ground truth with potential additional relevant items (i.e. documents and snippets), answer elements, and/or synonyms, which is still in progress. The online results pages for Phase A\footnote{\url{https://participants-area.bioasq.org/results/13b/phaseA/}}, Phase A+\footnote{\url{https://participants-area.bioasq.org/results/13b/phaseAplus/}}, and Phase B\footnote{\url{https://participants-area.bioasq.org/results/13b/phaseB/}} will be updated with the final results when available.

The overall performance of the participating systems in document and snippet retrieval (Phase A) per batch of task 13b is presented in Table {\ref{tab:bA_eval}}. Both the average and the top performance of the systems seem to drop in the last batch, indicating that the questions in this batch are more challenging for the systems. This could be related to the composition of this batch, which included more questions developed by new BioASQ experts, who have not contributed significantly to the development of the training dataset.  

\begin{table*}[!htb]
\caption{The evaluation measures for \textit{task 13b} per response type and question type~\cite{malakasiotis2020evaluation}.}\label{tab:b_eval}
\centering
\begin{tabular}{L{0.26\linewidth}M{0.21\linewidth}L{0.5\linewidth}}\hline
\textbf{Resp. type (Phase) }               & \textbf{Quest. type} & \textbf{Official measure }                                           \\\hline
Documents  (A)                         & All           & Mean Average Precision (MAP)                                \\\hline
Snippets (A)                            & All           & F1 (based on character overlaps)                     \\\hline
  & List          & F1                                                   \\
Exact ans. (A+ \& B)                                       & Yesno         & macro F1 on ``yes'' \& ``no'' classes               \\
                                       & Factoid       & Mean Reciprocal Rank (MRR)                                  \\\hline
Ideal ans. (A+ \& B)                  & All           & Manual scores for precision, recall, repetition, readability 
\\\hline 
\\
\end{tabular}
\caption{The average and top scores of participant systems in Phase A, task 13b. 
 }\label{tab:bA_eval}
    \centering
    \begin{tabular}{M{0.15\linewidth}M{0.2\linewidth}M{0.15\linewidth}M{0.2\linewidth}M{0.15\linewidth}}
    \hline
      & \multicolumn{2}{l}{\textbf{Documents}} & \multicolumn{2}{l}{\textbf{Snippets}} \\\hline
\textbf{Batch }& \textbf{Average MAP}     & \textbf{Top MAP}     & \textbf{Average F1}      & \textbf{Top F1}     \\\hline
1     & 0.231           & 0.425       & 0.053           & 0.120      \\
2     & 0.283           & 0.442       & 0.084           & 0.179      \\
3     & 0.175           & 0.324       & 0.052           & 0.110      \\
4     & 0.072           & 0.180       & 0.024           & 0.079     \\
     \hline                 
    \end{tabular}
\label{tab:bA_res_doc}
\end{table*}

\begin{figure}[!htbp]
\centerline{\includegraphics[width=1\textwidth]{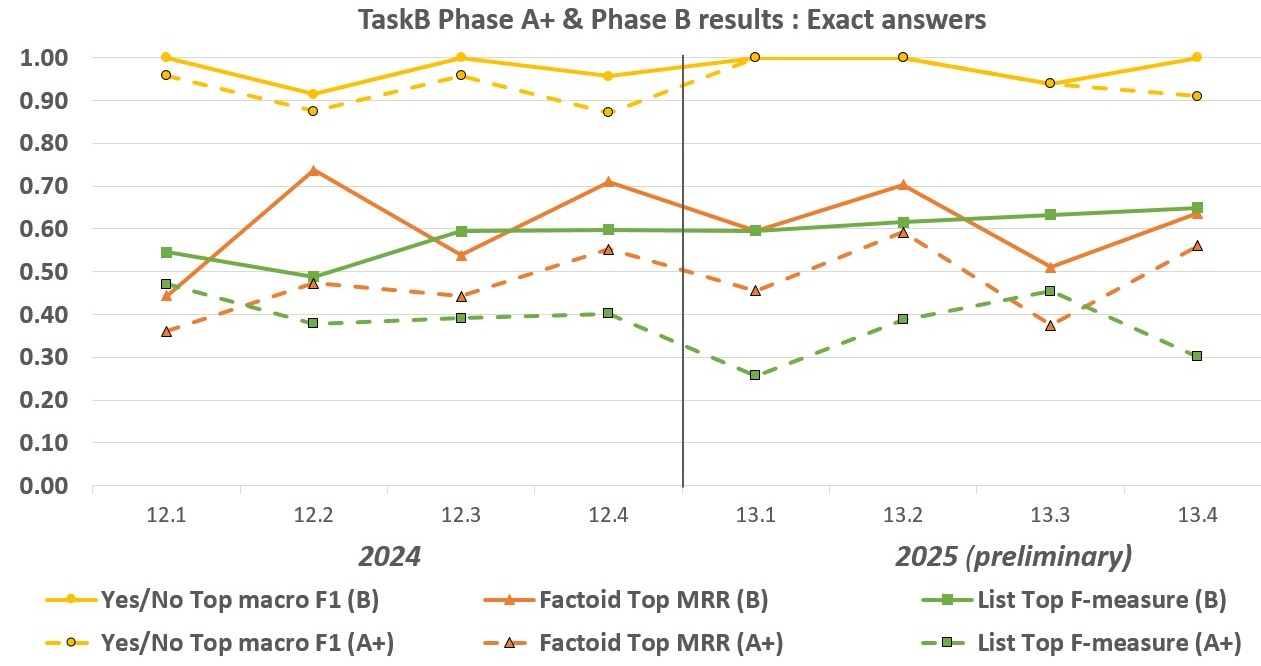}}
\caption{
The scores of the top systems in \textit{exact answer} generation, for Phase A+ (dashed lines) and B (solid lines), across the test sets of \textit{task 13b} and \textit{task 12b}~\cite{nentidis2024bioasq}. 
}\label{fig:Exact}
\end{figure}

The top performance of the participating systems in \textit{exact answer} generation per question type is presented in Figure {\ref{fig:Exact}} for both Phase A+ and Phase B of task 13b, in comparison to the respective performance in the previous version (12b).
These preliminary 13b results for phase B suggest that the top systems achieved scores comparable or higher to those of 12b in answering all types of questions (solid lines). 
In Phase A+ (dashed lines), the top performance is lower, as expected; however, for yesno questions in particular, it is very close to those of Phase B, revealing the increased capability of LLM-based models to address these questions even without being provided ground-truth relevant documents and snippets.
These results probably underestimate the performance of the top 13b systems in factoid and list questions, as the preliminary ground truth may miss some synonyms or alternative terms submitted by the participants. Such synonyms will be detected during the enrichment process and will be considered for the final results. 

\subsection{Task Synergy 13}

In \textit{task Synergy 13} we use the same evaluation measures described for \textit{task 13b}, considering only new material for the information retrieval part, an approach known as \textit{residual collection evaluation}~\cite{Salton1990}. 
In addition, due to the developing nature of the topics, no answer is available for all of the open questions in each round. Therefore, only the questions indicated as ``answer ready'' were evaluated for \textit{exact} and \textit{ideal answers} per round.

Table~\ref{tab:syn_data} presents the top performance achieved by participating systems per round, in task Synergy 13, for all types of responses.
During the four rounds of Synergy 13, the systems managed to identify enough relevant material to provide an answer to 59 of the 74 questions (about 80\%). In addition, they also managed to provide at least one \textit{ideal answer} which was considered of ground-truth quality by the respective expert for 35 questions (about 47\%).    
Overall, this dialogue between question-answering systems and biomedical experts allowed the progressive gathering of relevant documents and snippets and the generation of \textit{exact} and \textit{ideal answers} for open questions on developing topics, such as infectious, rare, and genetic diseases, and women’s and reproductive health.   

\begin{table}[!htb]
	\caption{The number of ``Answer Ready'' (AR) questions and the top system performance per round (R) in Task Synergy 13. Retrieval of documents (Top MAP) and snippets (Top F1). Generation of exact factoid (Top MRR), list (Top F1), and yesno (Top ma-F1) answers.
 }\label{tab:syn_data}
    \centering
    \begin{tabular}{M{0.04\linewidth}M{0.06\linewidth}M{0.14\linewidth}M{0.18\linewidth}M{0.14\linewidth}M{0.17\linewidth}M{0.19\linewidth}}
    \hline
\textbf{R} & \textbf{AR}	     & \textbf{Top MAP} 	     & \textbf{Top F1 Snip.} & \textbf{Top MRR} 	  & \textbf{Top F1 list}	& \textbf{Top macro-F1} \\\hline
1     & 19	&  	0.41		&	0.31   		&  	0.67		&    0.09	&      1       \\
2    & 33	&  	0.41		&	0.29   		&  	0.43		&     0.25	&       1      \\
3   & 49	&  	0.46		&	0.15   		&  	  0.5		&      0.26	&        1     \\
4   & 55	&  	0.47		&	0.25   		&  	 0.45		&     0.35	&         1   \\

     \hline                 
    \end{tabular}
\end{table}

\subsection{Task MultiClinSum}
The automatic evaluation of the MultiClinSum results was performed using both BERTScore \cite{zhang2020bertscore} and ROUGE-LSum metrics. Given the abstractive nature of the task, BERTScore was prioritized as the primary metric due to its superior capacity to capture semantic similarity between generated and reference summaries by leveraging contextualized embeddings, thus effectively recognizing meaning-preserving paraphrases and diverse lexical choices \cite{zhang2020bertscore}. 

In contrast, ROUGE-LSum—a sentence-level variant of the ROUGE metric \cite{Lin2004}—provides informative measures of summary quality by assessing the overlap of longest common subsequences between candidate and reference summaries. This offers valuable insights into the coverage and faithfulness of the generated content with respect to the original text. 

While ROUGE-LSum remains limited by its reliance on surface-level n-gram matching, its inclusion alongside BERTScore ensures a complementary perspective on summary quality, balancing semantic and lexical overlap considerations. The latter, however, was the prioritized metric for submission ranking purposes.

\begin{table}[!htb]
    \caption{Results of the MultiClinSum for each sub-track. Only the top-2 best teams are presented. The best result is in bold.}
    \label{tab:multiclinsum_res}
    \centering
    \renewcommand{\arraystretch}{1.1}
    \begin{tabular}{M{0.22\linewidth} M{0.14\linewidth} *{6}{M{0.09\linewidth}}}
        \hline
        \multicolumn{2}{c}{} & 
        \multicolumn{3}{c}{\textbf{BERTScore}} & 
        \multicolumn{3}{c}{\textbf{ROUGELSum}} \\
        \cline{3-8}
        \textbf{Team Name} & \textbf{Subtrack} & 
        \textbf{P} & \textbf{R} & \textbf{F1} & 
        \textbf{P} & \textbf{R} & \textbf{F1} \\
        \hline
        seemdog & English &  0.8795 & \textbf{0.8608} & \textbf{0.8698} & 0.3404 & \textbf{0.2398} & 0.267 \\
        pjmath.~\cite{pjmathematician_team} & English & \textbf{0.8821} & 0.8466 & 0.8637 & \textbf{0.4077} & 0.2343 & \textbf{0.2805} \\
        ggrazhdans~\cite{grazhdanski_team} & Spanish & \textbf{0.7699} & \textbf{0.747} & \textbf{0.7578} & 0.3639 & 0.2667 & 0.2899 \\
        pjmath.~\cite{pjmathematician_team} & Spanish & 0.7675 & 0.7392 & 0.7525 & \textbf{0.3684} & \textbf{0.2703} & \textbf{0.292} \\
        pjmath.~\cite{pjmathematician_team} & French & \textbf{0.7692} & \textbf{0.7459} & \textbf{0.7567} & \textbf{0.3481} & 0.2684 & \textbf{0.2843} \\
        BU team & French & 0.7248 & 0.7396 & 0.7315 & 0.2415 & \textbf{0.289} & 0.2466 \\
        pjmath.~\cite{pjmathematician_team} & Portuguese & \textbf{0.7644} & \textbf{0.7377} & \textbf{0.7502} & \textbf{0.35} & \textbf{0.2605} & \textbf{0.2803} \\
        ETS-PUCPR~\cite{elisaTerumi_team} & Portuguese & 0.7403 & 0.7351 & 0.737 & 0.2802 & 0.250 & 0.249 \\
        \hline
    \end{tabular}
\end{table}


BERTScore results in Table \ref{tab:multiclinsum_res} for non-english languages is impaired by the fact that a multilingual bert model is used instead of a language specific encoder model. Participants were able to submit up to 5 runs, the best of which was selected for each subtrack. There was a total o f 15, 14, 9 and 7 runs for English, Spanish, French and Portuguese respectively. 


\subsection{Task BioNNE-L}

Following prior research on entity linking~\cite{NEREL-BIO-COLING-2024,sapbert,GEBERT-CLEF-2023,BERGAMOT}, we address BioNNE-L as a retrieval task: given a mention, a model must retrieve the top-k concepts from the given UMLS dictionary and employ two ranking-based  evaluation metrics: (i) Accuracy@k and (ii) Mean Reciprocal Rank ($MRR$). Accuracy@k: Accuracy@k=1 if the correct UMLS CUI is retrieved at rank $\leq k$, and Accuracy@k=0 otherwise. $MRR = \frac{1}{|E|} \sum_{e \in E} \frac{1}{rank_{e}}$,  where $E$ is the set of entities, $|E|$ is the number of entities, $rank_{e}$ is the rank of entity $e$'s the first correctly retrieved concept among the top $k$ retrieved concepts. As baseline, we adopt zero-shot ranking using BERGAMOT~\cite{BERGAMOT} with each entity type processed independently to reduce memory footprint caused by extensive dictionary.

\begin{table}[!ht]
\centering
\caption{Official evaluation results of the BioNNE-L task for the multilingual and monolingual tracks in terms of Accuracy@1 (\textbf{@1}), Accuracy@5 (\textbf{@5}), and MRR. The best results for each track and metric are highlighted in \textbf{bold}.}
\label{tab:bionnel_results_mult}
\begin{tabular}{l|cccc|cccc|cccc}
\hline
\textbf{Team} & \multicolumn{4}{c}{\textbf{Multilingual}} & \multicolumn{4}{c}{\textbf{English}} & \multicolumn{4}{c}{\textbf{Russian}} \\
& \textbf{\#} & \textbf{@1} & \textbf{@5 } & \textbf{MRR} & \textbf{\#} & \textbf{@1} & \textbf{@5 } & \textbf{MRR} & \textbf{\#} & \textbf{@1} & \textbf{@5 } & \textbf{MRR} \\ 
\cmidrule{2-5} \cmidrule{6-9} \cmidrule{10-13}
 verbanexialab & --- & --- & --- & --- & 1 & \textbf{0.70} & 0.80 & \textbf{0.74}  & --- & --- & --- & --- \\
 LYX\_DMIIP\_FDU & 1 & \textbf{0.68}  & \textbf{0.84}  & \textbf{0.75} & 2 & 0.66 & \textbf{0.84} & \textbf{0.74}  & 2 & 0.71 & \textbf{0.84} & \textbf{0.76} \\
 BlancaPlanca & 2 & 0.67  & 0.81  & 0.73 & 3 & 0.64 & 0.83 & 0.72  & 1 & \textbf{0.72} & 0.83 & \textbf{0.76} \\
 MSM Lab  & 3 & 0.63  & 0.76  & 0.69  & 4 & 0.64 & 0.82 & 0.71  & 4 & 0.65 & 0.74 & 0.69 \\
 dstepakov & 4 & 0.63  & 0.71  & 0.66  & --- & --- & --- & ---  & 3 & 0.70 & 0.76 & 0.72 \\
 ICUE & 5 & 0.58  & 0.76  & 0.66  & 6 & 0.51 & 0.79 & 0.62  & 5 & 0.62 & 0.72 & 0.67 \\
 baseline & 6 & 0.53  & 0.70  & 0.60  & 5 & 0.57 & 0.78 & 0.66 &  6 & 0.52 & 0.59 & 0.55 \\
 NLPIMP & 7 & 0.41  & 0.58  & 0.48  & --- & --- & --- & --- & --- & --- & --- & ---   \\

\hline

\hline
\end{tabular}
\end{table}

The official evaluation results, ordered by Accuracy@1 value, for BioNNE-L are summarized in Table~\ref{tab:bionnel_results_mult}. Most of the participants adopted various BERT-based~\cite{Devlin2018} with the top-performance achieved by domain-specific models, such as BERGAMOT, SapBERT. Specifically, Team  LYX\_DMIIP\_FDU ranked the first in the multilingual track and the second rank in the two monolingual track by fine-tuning BERGAMOT. Top 1 results for the Russian and English data are achieved by multilingual BERGAMOT (Team BlancaPlanca) and English SapBERT (Team verbanexialab) models, respectively.

\subsection{Task ELCardioCC}

The results of the participants for the ELCardioCC task are presented in tables \ref{tab:elcard-ner}, \ref{tab:elcard-el} and \ref{tab:elcard-mlc-x}. For each subtask, the table displays one system per team that achieved the highest F1-score.  

\begin{table}[!ht]
\caption{Performance of participating systems in the ELCardioCC Named Entity Recognition (NER) subtask. Results are reported using micro-averaged precision, recall, and F1 score.}
\centering
\begin{tabular}{|c|l|c|c|c|}
\hline
\textbf{Team} & \textbf{System} & \textbf{Recall} & \textbf{Precision} & \textbf{Micro-F1} \\
\hline
\multirow{1}{*}{bhuang} 
  & 5nm & 0.6448 & 0.5205 & 0.5761 \\
\hline
\multirow{1}{*}{droidlyx} 
  & system1 & \textbf{0.7059} & \textbf{0.7618} & \textbf{0.7328} \\
\hline
\multirow{1}{*}{ELCardioCC\_baseline} 
  & mbert\_baseline & 0.6959 & 0.7460 & 0.7201 \\
\hline
\multirow{1}{*}{pjmathematician} 
  & config1 & 0.2484 & 0.2586 & 0.2534 \\
\hline
\multirow{1}{*}{enigma} 
  & greek-bert-exact-bge-m3 & 0.7012 & 0.7328 & 0.7167 \\
\hline 
\end{tabular}
\label{tab:elcard-ner}
\end{table}

\begin{table}[!ht]
\caption{Performance of participating systems in the ELCardioCC Entity Linking (EL) subtask. Results are reported using micro-averaged precision, recall, and F1 score.}
\centering
\begin{tabular}{|c|l|c|c|c|}
\hline
\textbf{Team} & \textbf{System} & \textbf{Recall} & \textbf{Precision} & \textbf{Micro-F1} \\
\hline
\multirow{1}{*}{bhuang} 
  & 5nm & 0.5927 & 0.4852 & 0.5336 \\
\hline
\multirow{1}{*}{droidlyx} 
  & system1 & 0.6529 & \textbf{0.7046} & \textbf{0.6778} \\
\hline
\multirow{1}{*}{ELCardioCC\_baseline} 
  & EL\_baseline & 0.6476 & 0.6942 & 0.6701 \\
\hline
\multirow{1}{*}{pjmathematician} 
  & config1 & 0.0616 & 0.0642 & 0.0629 \\
\hline
\multirow{1}{*}{enigma} 
  & greek-bert-exact-bge-m3 & \textbf{0.6548} & 0.6844 & 0.6693 \\
\hline
\end{tabular}
\label{tab:elcard-el}
\end{table}

\begin{table}[ht!]
\caption{Performance of participating systems in ELCardioCC Subtask 3a (Multi-label Classification) and Subtask 3b (Explainable AI). Metrics include Precision (P), Recall (R), and Micro-F1 score. A dash (–) indicates that the system did not participate in the corresponding subtask.}
\centering
\begin{tabular}{|c|l|ccc|ccc|}
\hline
\textbf{Team} & \textbf{System} 
& \multicolumn{3}{c|}{\textbf{Subtask 3a (MLC)}} 
& \multicolumn{3}{c|}{\textbf{Subtask 3b (X)}} \\
\cline{3-8}
& & \textbf{P} & \textbf{R} & \textbf{F1} & \textbf{P} & \textbf{R} & \textbf{F1} \\
\hline
\multirow{2}{*}{ELCardioCC\_baseline} 
  & MLCX1\_baseline & 0.9339 & 0.7422 & 0.8271 & - & - & - \\
  & MLCX2\_baseline & \textbf{0.9531} & 0.5864 & 0.7261 & \textbf{0.6050} & \textbf{0.4442} & \textbf{0.5122 }\\
\hline
\multirow{1}{*}{bhuang} 
  & 1nm & 0.6205 & 0.7676 & 0.6863 & - & - & - \\
\hline
\multirow{1}{*}{droidlyx} 
  & system1 & 0.8569 & 0.8377 & \textbf{0.8472} & - & - & - \\
\hline
\multirow{1}{*}{kbogas} 
  & w2l\_cb & 0.2115 & 0.3421 & 0.2614 & - & - & - \\
\hline
\multirow{2}{*}{pjmathematician} 
  & config4 & 0.6056 & 0.2257 & 0.3288 & 0.2326 & 0.0932 & 0.1331 \\
  & config5 & 0.5860 & 0.2656 & 0.3655 & - & - & - \\
\hline
\end{tabular}
\label{tab:elcard-mlc-x}
\end{table}

The team droidlyx \cite{bionnel-LYX-DMIIP-FDU} consistently achieved the highest micro-F1 scores across all subtasks, leading NER (0.733), EL (0.678), and performing strongly in MLC-X (0.847). The ELCardioCC baseline remained highly competitive, particularly excelling in MLC-X with a micro-F1 of 0.827 and maintaining strong performance in NER and EL. Enigma's systems \cite{enigma-elcardiocc} were consistently near the top in NER and EL but did not participate in MLC-X, while bhuang \cite{MultilingualLLMBasedMultiStateSystem} showed promising results especially in classification, though with more variability. Finally, the pjmathematician \cite{multi-embedding-prompt-driven} system demonstrated significantly low performance . 
\subsection{Task GutBrainIE}
Submitted runs were evaluated using micro- and macro-averaged precision, recall, and F1-score, with micro-F1 used as the reference measure for the leaderboards since it is better suited when classes are imbalanced. 

We adopted a baseline employing a fine-tuned NuNER model for NER \cite{bogdanov2024nuner} and a fine-tuned ATLOP model for all RE subtasks \cite{zhou2021atlop}. The baseline has also been used to annotate the bronze collection automatically.

Tables \ref{tab:gutbrainie_T61_scores}-\ref{tab:gutbrainie_T623_scores} show each team’s top run beating the baseline for NER, TB-RE, TT-RE, and TM-RE, respectively. 
The performance gap between the subtasks is noticeable. 
While NER obtained a top micro-F1 of 0.84 utilizing pretrained biomedical transformers, RE subtasks were more challenging: both TB-RE and TT-RE peaked at approximately 0.65-0.69 micro-F1, and TM-RE reached only 0.46 micro-F1, demonstrating the added difficulty of simultaneously locating and labeling entities and identifying relations among these. 
\begin{table}[!htb]
    \centering
    \caption{Performance metrics of each team’s top run beating the baseline for NER. The best result is in bold, the second-best is underlined (micro-averaged).}
    \label{tab:gutbrainie_T61_scores}
    \begin{tabular}{L{0.27\linewidth}L{0.3\linewidth}M{0.15\linewidth}M{0.1\linewidth}M{0.1\linewidth}}
        \hline
        \textbf{Team ID} & \textbf{Run name} & \textbf{ Precision } & \textbf{ Recall } & \textbf{  F1  } \\
        \hline
        GutUZH \cite{team_GutUZH_paper_gbie} & AugEnsemble & \textbf{0.8384} & 0.8432 & \textbf{0.8408} \\
        Gut-Instincts \cite{team_Gut-Instincts_paper_gbie} & 5eedev & 0.8286 & 0.8480 & \underline{0.8382} \\
        NLPatVCU \cite{team_NLPatVCU_paper_gbie} & ensemble1 & 0.8255 & \underline{0.8488} & 0.8370 \\
        ICUE \cite{team_ICUE_paper_gbie} & ensemble5-th10 & \underline{0.8369} & 0.8294 & 0.8331 \\
        LYX-DMIIP-FDU \cite{bionnel-LYX-DMIIP-FDU} & EnsembleBERT & 0.8020 & \textbf{0.8513} & 0.8259 \\
        ata2425ds [NA] & transformer & 0.7914 & 0.8432 & 0.8164 \\
        greenday \cite{team_greenday_paper_gbie} & llmner & 0.7957 & 0.8278 & 0.8114 \\
        Graphswise-1 \cite{team_graphwise-1_paper_gbie} & NERWise & 0.8066 & 0.7955 & 0.8010 \\\hline
        BASELINE \cite{BioASQ2025taskGutBrainIE} & NuNerZero-Finetuned & 0.7639 & 0.8238 & 0.7927 \\
        \hline
    \end{tabular}
\end{table}

\begin{table}[!htb]
    \caption{Performance metrics of each team’s top run beating the baseline for TB-RE. The best result is in bold, the second-best is underlined (micro-averaged).}
    \label{tab:gutbrainie_T621_scores}    
    \centering
    \begin{tabular}{L{0.25\linewidth}L{0.3\linewidth}M{0.15\linewidth}M{0.1\linewidth}M{0.1\linewidth}}
        \hline
        \textbf{Team ID} & \textbf{Run name} & \textbf{ Precision } & \textbf{ Recall } & \textbf{  F1  } \\
        \hline
        Gut-Instincts \cite{team_Gut-Instincts_paper_gbie} & 6219eedev3re & 0.6304 & \textbf{0.7532} & \textbf{0.6864} \\
        ONTUG \cite{team_ONTUG_paper_gbie} & ElectraCLEANR & 0.7121	& 0.6104 & \underline{0.6573} \\
        Graphswise-1 \cite{team_graphwise-1_paper_gbie} & AtlopOnto & 0.7418 & 0.5844	& 0.6538 \\
        ataupd2425-pam \cite{team_ataupd2425-pam_paper_gbie} & BiomedNLP-FULL\_DATASET & 0.5671 & \underline{0.7316} & 0.6389 \\
        BIU-ONLP \cite{team_BIU-ONLP_paper_gbie} & RobertaLarge & \underline{0.7453} & 0.5195 & 0.6122 \\ \hline
        BASELINE \cite{BioASQ2025taskGutBrainIE} & Atlop-Finetuned & \textbf{0.7584}	& 0.4892 & 0.5947 \\
        \hline
    \end{tabular}
\end{table}

\begin{table}[!htb]
    \caption{Performance metrics of each team’s top run beating the baseline for TT-RE. The best result is in bold, the second-best is underlined (micro-averaged).}
    \label{tab:gutbrainie_T622_scores}    
    \centering
    \begin{tabular}{L{0.25\linewidth}L{0.3\linewidth}M{0.15\linewidth}M{0.1\linewidth}M{0.1\linewidth}}
        \hline
        \textbf{Team ID} & \textbf{Run name} & \textbf{ Precision } & \textbf{ Recall } & \textbf{  F1  } \\
        \hline
        Gut-Instincts \cite{team_Gut-Instincts_paper_gbie} & 6229eedev3re & 0.6280 & \underline{0.7572} & \textbf{0.6866} \\
        ataupd2425-pam \cite{team_ataupd2425-pam_paper_gbie} & BiomedNLP-FULL\_DATASET & 0.5853 & 0.7202 & \underline{0.6458} \\
        ONTUG \cite{team_ONTUG_paper_gbie} & ElectraCLEANR & 0.7059 & 0.5926 & 0.6443 \\
        Graphswise-1 \cite{team_graphwise-1_paper_gbie} & AtlopOnto & 0.7326 & 0.5638 & 0.6372 \\
        ICUE \cite{team_ICUE_paper_gbie} & biolinkbertl\_pp & 0.4974 & \textbf{0.7860} & 0.6093 \\
        BIU-ONLP \cite{team_BIU-ONLP_paper_gbie} & RobertaLarge & \underline{0.7362} & 0.4938 & 0.5911 \\ \hline
        BASELINE \cite{BioASQ2025taskGutBrainIE} & Atlop-Finetuned & \textbf{0.7533} & 0.4650 & 0.5751 \\
        \hline
    \end{tabular}
\end{table}

\begin{table}[!htb]
    \caption{Performance metrics of each team’s top run beating the baseline for TM-RE. The best result is in bold, the second-best is underlined (micro-averaged).}
    \label{tab:gutbrainie_T623_scores}    
    \centering
    \begin{tabular}{L{0.25\linewidth}L{0.3\linewidth}M{0.15\linewidth}M{0.1\linewidth}M{0.1\linewidth}}
        \hline
        \textbf{Team ID} & \textbf{Run name} & \textbf{ Precision } & \textbf{ Recall } & \textbf{  F1  } \\ 
        \hline
        Gut-Instincts \cite{team_Gut-Instincts_paper_gbie} & 6239eedev3re & 0.4215 & \textbf{0.5147} & \textbf{0.4635} \\ 
        Graphswise-1 \cite{team_graphwise-1_paper_gbie} & AtlopOnto & \underline{0.4686} & 0.3097 & \underline{0.3729} \\ 
        ICUE \cite{team_ICUE_paper_gbie} & biolinkbertl\_pp & 0.2858 & \underline{0.5054} & 0.3651 \\ 
        LYX-DMIIP-FDU \cite{bionnel-LYX-DMIIP-FDU} & BioLinkBERT & 0.3682 & 0.3257 & 0.3457 \\ 
        ONTUG \cite{team_ONTUG_paper_gbie} & ElectraCLEANR & 0.3529 & 0.3231 & 0.3373 \\ \hline
        BASELINE \cite{BioASQ2025taskGutBrainIE} & Atlop-Finetuned & \textbf{0.4986} & 0.2453 & 0.3288 \\ \hline
    \end{tabular}
\end{table}

\section{Conclusions}
\label{sec:conclusion}

This paper provides an overview of the thirteenth BioASQ challenge.
This year, BioASQ consisted of six tasks: 
(i) \textit{Task 13b} on biomedical semantic question answering. 
(ii) \textit{Task Synergy13} on question answering for developing biomedical topics.
(iii) \textit{Task MultiClinSum} on multilingual clinical summarization. 
(iv) \textit{Task BioNNE-L} on nested named entity linking in Russian and
English. 
(v) \textit{Task ELCardioCC} on clinical coding in cardiology.
(vi) \textit{Task GutBrainIE} on gut-brain interplay information extraction.

The results for Task 13b suggest that the top participant systems achieved high scores, especially for yes/no answer generation, even in Phase A+, where no ground-truth relevant material was given. For list and factoid questions, system performance is less consistent, especially in Phase A+, indicating the presence of room for improvement. For these questions, the availability of ground-truth relevant material seems to allow the systems to provide answers of better quality.
This highlights the importance of Phase A, on the automated retrieval of relevant material, where the top performance is less consistent across batches, potentially affected by the domain of the expert posing the questions. 
A diverse set of retrieval and generation techniques was applied, including traditional methods, LLM-based frameworks, and integration of domain-specific knowledge.
The results of task Synergy13, aligned with those of previous versions, suggest that state-of-the-art systems can be a useful tool for biomedical scientists in need of specialized information for developing problems, despite their limitations and room for improvement.

The new task MultiClinSum presented new challenging sub-tasks about text summarization of clinical case reports in Spanish, English, French and Portuguese. 
This task introduces the nuance of creating clinical Text Summarization systems specifically for the cardiology domain. In addition, it expands the range of the task beyond Spanish by introducing a sub-track that also involves English and Italian text. 
The results highlight the importance of having data specific to the language and specialty the systems are going to be applied in, even within domains that are already quite specific, like the clinical one. 

The BioNNE-L task focused on the linking of biomedical entities for disorders, chemicals, and anatomical structures in Russian and English texts, addressing challenges such as nested entities and cross-language linking amid incomplete low-resource vocabularies. Despite the overall prevalence of LLMs in numerous domains and tasks, the top-performing systems for BioNNE-L utilized biomedical BERT-based retrieval and reranking architectures, highlighting the importance of task and domain-specific methods for information extraction.

The ELCardioCC task centered on extracting and classifying medical entities from Greek discharge letters, drawing participation from five teams who submitted a diverse set of systems across three subtasks. Most approaches leveraged transformer-based models—particularly BERT variants and multilingual LLMs—with strategies ranging from fine-tuned token classification to prompt-based extraction and embedding-based entity linking. The MLC-X subtask saw innovative uses of multilingual embeddings and LLM reasoning to predict and justify ICD-10 codes. Baseline models, built on multilingual BERT architectures, offered simple yet effective benchmarks for each subtask, emphasizing clarity and reproducibility.

The GutBrainIE task, centered on information extraction for the gut-brain axis, challenged participants with Named Entity Recognition and increasingly fine-grained Relation Extraction subtasks.
Teams achieving the strongest performance employed supervised deep-learning strategies, combining pretrained biomedical language models with ensemble strategies. 
Only a few participants experimented with prompt-based or generative approaches; however, these generally obtained lower scores, confirming the need to develop specialized models to effectively extract complex entities and relations in a specific biomedical domain.  

Overall, the participation in BioASQ 13 was significantly increased, both due to increased interest in the new versions of its already established tasks, as well as due to the introduction of four novel tasks. 
Several participating systems achieved competitive performance on the BioASQ tasks, and some of them managed to improve over the baselines or the state-of-the-art performance from previous years.
Aligned with previous versions, BioASQ keeps pushing the research frontier in biomedical semantic indexing and question answering for thirteen years now, offering both well-established and new tasks. 
Initially, it extended beyond the English language and biomedical literature with the introduction of the task MESINESP \cite{luis2020overview} and continued consistently ever since. In this thirteenth edition, BioASQ was further extended with four new tasks, MultiClinSum~\cite{BioASQ2025MultiClinSum}, BioNNE-L~\cite{bionnel-overview-2025}, ElCardioCC~\cite{BioASQ2025ElCardioCC}, and GrutBrainIE~\cite{BioASQ2025taskGutBrainIE}. 
As a result, BioASQ 13 offered tasks in six languages (English, Spanish, French, Portuguese, Russian, and Greek), three types of documents (biomedical articles, clinical case reports, and discharge letters), and two specialized domains within biomedicine (cardiology and gut-brain interaction).

The future directions for the BioASQ challenge involve further expanding the benchmark dataset for question answering through a community-driven approach, broadening the network of biomedical experts participating in the Synergy task, and enhancing the scope of resources used in the BioASQ tasks. This includes incorporating additional document types, multiple languages, and more specialized sub-domains within biomedicine.

\section{Acknowledgments}
The thirteenth edition of BioASQ is sponsored by Ovid, Atypon Systems Inc, and Elsevier.
The MEDLINE/PubMed data resources considered in this work were accessed courtesy of the U.S. National Library of Medicine.
BioASQ is grateful to the CMU team for providing the \textit{exact answer} baselines for task 13b.
This research was funded by the Ministerio de Ciencia e Innovación (MICINN) under project BARITONE (TED2021-129974B-C22). This work is also supported by the European Union’s Horizon Europe Co-ordination \& Support Action under Grant Agreement No 101080430 (AI4HF), as well as Grant Agreement No 101057849 (DataTool4Heartproject).
The work on the BioNNE-L task was supported by the Russian Science Foundation [grant number 23-11-00358].
ELCardioCC has been partially supported by project MIS 5154714 of the National Recovery and Resilience Plan Greece 2.0 funded by the European Union under the NextGenerationEU Program.
The work on the GutBrainIE task was supported by the HEREDITARY Project, as part of the European Union's Horizon Europe research and innovation programme (GA 101137074).
%
%
%
\bibliographystyle{splncs04}
\bibliography{BioASQ13.bib}

\begin{thebibliography}{10}
\providecommand{\url}[1]{\texttt{#1}}
\providecommand{\urlprefix}{URL }
\providecommand{\doi}[1]{https://doi.org/#1}

\bibitem{team_Gut-Instincts_paper_gbie}
Andersen, L.R., Gardshodn, M.I., Dolmer, M.H., Rodriguez, J.M., Dell'Aglio, D.: {Trusting Gut Instincts: Transformer-Based Extraction of Structured Data from Gut-Brain Axis Publications}. In: Faggioli, G., Ferro, N., Rosso, P., Spina, D. (eds.) CLEF 2025 Working Notes (2025)

\bibitem{Angulo25}
Angulo, J., Yeste, V.: {AQAMS and AQAMS2: Multi Agent Systems for Biomedical Question Answering }. In: Faggioli, G., Ferro, N., Rosso, P., Spina, D. (eds.) CLEF 2025 Working Notes (2025)

\bibitem{appleton2018gut}
Appleton, J.: The gut-brain axis: influence of microbiota on mood and mental health. Integrative Medicine: A Clinician's Journal  \textbf{17}(4), ~28 (2018)

\bibitem{Ateia25}
Ateia, S., Kruschwitz, U.: {Can Language Models Critique Themselves? Investigating Self-Feedback for Retrieval Augmented Generation at BioASQ 2025 }. In: Faggioli, G., Ferro, N., Rosso, P., Spina, D. (eds.) CLEF 2025 Working Notes (2025)

\bibitem{scibert}
Beltagy, I., Lo, K., Cohan, A.: {SciBERT: Pretrained Language Model for Scientific Text}. In: EMNLP (2019)

\bibitem{Chen25}
Bing-Chen, C., Han, J.C., Hung, H.C., Tsai, R.T.H.: {NCU-IISR: Biomedical Question Answering via Gemini and GPT APIs in the BioASQ 13b Phase B Challenge }. In: Faggioli, G., Ferro, N., Rosso, P., Spina, D. (eds.) CLEF 2025 Working Notes (2025)

\bibitem{bodenreider2004unified}
Bodenreider, O.: {The unified medical language system (UMLS): integrating biomedical terminology}. Nucleic acids research  \textbf{32}(suppl\_1),  D267--D270 (2004)

\bibitem{bogdanov2024nuner}
Bogdanov, S., Constantin, A., Bernard, T., Crabbé, B., Bernard, E.: {NuNER: Entity Recognition Encoder Pre-training via LLM-Annotated Data} (2024)

\bibitem{Borazio25}
Borazio, F., Croce, D., Basili, R.: {UniTor at BioASQ 2025: Modular Biomedical QA with Synthetic Snippets and Multiple Task Answer Generation }. In: Faggioli, G., Ferro, N., Rosso, P., Spina, D. (eds.) CLEF 2025 Working Notes (2025)

\bibitem{bionnel-burlova}
Burlova, A.: {Navigating Partial UMLS Terminology: GAT Embeddings and Confidence Analysis for Multilingual Concept Linking}. In: Faggioli, G., Ferro, N., Rosso, P., Spina, D. (eds.) CLEF 2025 Working Notes (2025)

\bibitem{carabotti2015gut}
Carabotti, M., Scirocco, A., Maselli, M.A., Severi, C.: The gut-brain axis: interactions between enteric microbiota, central and enteric nervous systems. Annals of gastroenterology: quarterly publication of the Hellenic Society of Gastroenterology  \textbf{28}(2), ~203 (2015)

\bibitem{clark2020electra}
Clark, K., Luong, M.T., Le, Q.V., Manning, C.D.: {ELECTRA: Pre-training Text Encoders as Discriminators Rather Than Generators}. In: International Conference on Learning Representations (2020)

\bibitem{team_lasigeBioTM_paper_gbie}
Concei{\c{c}}{\~a}o, S.I.R., Lopes, P.R.C., Couto, F.M.: {lasigeBioTM at BioASQ25 Task GutBrainIE - Lean Large language models with syntactic features}. In: Faggioli, G., Ferro, N., Rosso, P., Spina, D. (eds.) CLEF 2025 Working Notes (2025)

\bibitem{cryan2020gut}
Cryan, J.F., O'Riordan, K.J., Sandhu, K., Peterson, V., Dinan, T.G.: The gut microbiome in neurological disorders. The Lancet Neurology  \textbf{19}(2),  179--194 (2020)

\bibitem{bionnel-ICUE}
D.~Lain, A., Lee, C., Doneva, S.E., Rodr{\'\i}guez-Cubillos, M.J., Castagnari, E., Simpson, T.I., , Posma, J.M.: {Multilingual and Nested Biomedical Named Entity Normalisation via Candidate Retrieval and Lightweight Large Language Model Disambiguation}. In: Faggioli, G., Ferro, N., Rosso, P., Spina, D. (eds.) CLEF 2025 Working Notes (2025)

\bibitem{team_graphwise-1_paper_gbie}
Datseris, A., Kuzmanov, M., Nikolova-Koleva, I., Taskov, D., Boytcheva, S.: {Graphwise @ CLEF-2025 GutBrainIE: Towards Automated Discovery of Gut-Brain Interactions: Deep Learning for NER and Relation Extraction from PubMed Abstracts}. In: Faggioli, G., Ferro, N., Rosso, P., Spina, D. (eds.) CLEF 2025 Working Notes (2025)

\bibitem{Devlin2018}
Devlin, J., Chang, M.W., Lee, K., Toutanova, K.: {BERT}: Pre-training of deep bidirectional transformers for language understanding. In: Burstein, J., Doran, C., Solorio, T. (eds.) Proceedings of the 2019 Conference of the North {A}merican Chapter of the Association for Computational Linguistics: Human Language Technologies, Volume 1 (Long and Short Papers). pp. 4171--4186. ACL, Minneapolis, Minnesota (Jun 2019). \doi{10.18653/v1/N19-1423}

\bibitem{BioASQ2025ElCardioCC}
Dimitriadis, D., Patsiou, V., Stoikopoulou, E., Toumpas, A., Kipouros, A., Papadopoulos, D., Bekiaridou, A., Barmpagiannos, K., Vasilopoulou, A., Barmpagiannos, A., Samaras, A., Giannakoulas, G., Tsoumakas, G.: {Overview of ElCardioCC Task on Clinical Coding in Cardiology at BioASQ 2025}. In: Faggioli, G., Ferro, N., Rosso, P., Spina, D. (eds.) CLEF 2025 Working Notes (2025)

\bibitem{Duenas_Romero}
Dueñas~Romero, S., Ureña-López, L.A., Martínez-Cámara, E.: {SINAI at CLEF 2025: A Multi-Stage RAG Pipeline for Biomedical Semantic Question Answering }. In: Faggioli, G., Ferro, N., Rosso, P., Spina, D. (eds.) CLEF 2025 Working Notes (2025)

\bibitem{feng-etal-2022-language}
Feng, F., Yang, Y., Cer, D., Arivazhagan, N., Wang, W.: Language-agnostic {BERT} sentence embedding. In: Proceedings of the 60th Annual Meeting of the Association for Computational Linguistics (Volume 1: Long Papers). pp. 878--891. ACL, Dublin, Ireland (May 2022). \doi{10.18653/v1/2022.acl-long.62}

\bibitem{Galat25}
Galat, D., Molla-Aliod, D.: {LLM Ensemble for RAG: Role of context length in zero-shot Question Answering for BioASQ Challenge }. In: Faggioli, G., Ferro, N., Rosso, P., Spina, D. (eds.) CLEF 2025 Working Notes (2025)

\bibitem{luis2020overview}
Gasco, L., Nentidis, A., Krithara, A., Estrada-Zavala, D., Toshiyuki~Murasaki, R., Primo-Pe{\~n}a, E., Bojo-Canales, C., Paliouras, G., Krallinger, M.: {Overview of BioASQ 2021-MESINESP track. Evaluation of advance hierarchical classification techniques for scientific literature, patents and clinical trials.} In: Proceedings of the 9th BioASQ Workshop (2021)

\bibitem{ghaisas2016gut}
Ghaisas, S., Maher, J., Kanthasamy, A.: Gut microbiome in health and disease: Linking the microbiome--gut--brain axis and environmental factors in the pathogenesis of systemic and neurodegenerative diseases. Pharmacology \& therapeutics  \textbf{158},  52--62 (2016)

\bibitem{grazhdanski_team}
Grazhdanski, G.: Group relative policy optimization for spanish clinical case report summarization. In: Faggioli, G., Ferro, N., Rosso, P., Spina, D. (eds.) CLEF 2025 Working Notes (2025)

\bibitem{pubmedbert}
Gu, Y., Tinn, R., Cheng, H., Lucas, M., Usuyama, N., Liu, X., Naumann, T., Gao, J., Poon, H.: Domain-specific language model pretraining for biomedical natural language processing. ACM Transactions on Computing for Healthcare (HEALTH)  \textbf{3}(1),  1--23 (2021)

\bibitem{team_greenday_paper_gbie}
Gupta, H.P., Banerjee, R.: {LLMs for Biomedical NER}. In: Faggioli, G., Ferro, N., Rosso, P., Spina, D. (eds.) CLEF 2025 Working Notes (2025)

\bibitem{team_GutUZH_paper_gbie}
Han, J., Liu, Y.: {GutUZH at CLEF2025 BioASQ Task 6: a method of SOTA performance with the best results at GutBrainIE NER subtask 1}. In: Faggioli, G., Ferro, N., Rosso, P., Spina, D. (eds.) CLEF 2025 Working Notes (2025)

\bibitem{MultilingualLLMBasedMultiStateSystem}
Huang, B.: Clinical entity recognition and linking in greek discharge letters using multilingual-llm-based multi-stage system. In: Faggioli, G., Ferro, N., Rosso, P., Spina, D. (eds.) CLEF 2025 Working Notes (2025)

\bibitem{huguet-cabot-navigli-2021-rebel-relation}
Huguet~Cabot, P.L., Navigli, R.: {REBEL}: Relation extraction by end-to-end language generation. In: Findings of the Association for Computational Linguistics: EMNLP 2021. pp. 2370--2381. ACL, Punta Cana, Dominican Republic (Nov 2021), \url{https://aclanthology.org/2021.findings-emnlp.204}

\bibitem{Jonker25}
Jonker, R.A.A., Almeida, T., Almeida, J., Matos, S.: {BIT.UA at BioASQ 13B: Revisiting Evaluation, DPRF-Enhanced Retrieval and Fine-Tuned LLMs}. In: Faggioli, G., Ferro, N., Rosso, P., Spina, D. (eds.) CLEF 2025 Working Notes (2025)

\bibitem{team_ONTUG_paper_gbie}
Kantz, B., Waldert, P., Lengauer, S., Schreck, T.: {Constrained Linked Entity ANnotation using RAG (CLEANR)}. In: Faggioli, G., Ferro, N., Rosso, P., Spina, D. (eds.) CLEF 2025 Working Notes (2025)

\bibitem{team_BIU-ONLP_paper_gbie}
Keinan, R., Cohen, A.D.N., Tsarfaty, R.: {From Named Entities to Relations: End-to-End Biomedical Information Extraction}. In: Faggioli, G., Ferro, N., Rosso, P., Spina, D. (eds.) CLEF 2025 Working Notes (2025)

\bibitem{Kim25}
Kim, H., Lee, H., Cho, Y., Park, J., Park, J., Park, S., Chok, Y.T., Baek, S., Lee, D., Kang, J.: {Prompting Matters: Snippet-Aware Strategies for Biomedical QA with LLMs in BioASQ 13b }. In: Faggioli, G., Ferro, N., Rosso, P., Spina, D. (eds.) CLEF 2025 Working Notes (2025)

\bibitem{krithara2023bioasq}
Krithara, A., Nentidis, A., Bougiatiotis, K., Paliouras, G.: {BioASQ-QA: A manually curated corpus for Biomedical Question Answering}. Scientific Data  \textbf{10}(1), ~170 (2023)

\bibitem{Krithara2016overview}
Krithara, A., Nentidis, A., Paliouras, G., Kakadiaris, I.: {Results of the 4th edition of BioASQ Challenge}. In: Proceedings of the Fourth BioASQ workshop (2016), \url{https://www.aclweb.org/anthology/W16-3101.pdf}

\bibitem{krithara2021bioasq}
Krithara, A., Nentidis, A., Paliouras, G., Krallinger, M., Miranda, A.: {BioASQ at CLEF2021: large-scale biomedical semantic indexing and question answering}. In: Advances in Information Retrieval: ECIR 2021, Virtual Event, March 28--April 1, 2021, Proceedings, Part II 43. pp. 624--630. Springer (2021)

\bibitem{Synergy_jamia}
Krithara, A., Nentidis, A., Vandorou, E., Katsimpras, G., Almirantis, Y., Arnal, M., Bunevicius, A., Farre-Maduell, E., Kassiss, M., Konstantakos, V., Matis-Mitchell, S., Polychronopoulos, D., Rodriguez-Pascual, J., Samaras, E.G., Samiotaki, M., Sanoudou, D., Vozi, A., Paliouras, G.: {BioASQ Synergy: a dialogue between question-answering systems and biomedical experts for promoting COVID-19 research}. Journal of the American Medical Informatics Association p. ocae232 (08 2024). \doi{10.1093/jamia/ocae232}

\bibitem{team_ICUE_paper_gbie}
Lee, C., Doneva, S., Rodriguez-Cubillos, M., Castagnari, E., Lain, A., Posma, J., Simpson, T.I.: {Understanding Gut-Brain Interplay in Scientific Literature: A Hybrid Approach from Classification to Generative LLM Reasoning}. In: Faggioli, G., Ferro, N., Rosso, P., Spina, D. (eds.) CLEF 2025 Working Notes (2025)

\bibitem{lee2019biobert}
Lee, J., Yoon, W., Kim, S., Kim, D., Kim, S., So, C.H., Kang, J.: {BioBERT: a pre-trained biomedical language representation model for biomedical text mining}. Bioinformatics  \textbf{36}(4),  1234--1240 (09 2019). \doi{10.1093/bioinformatics/btz682}

\bibitem{bionnel-MSM-Lab}
Li, C., Zheng, X., Liu, S.: {BIBERT on Biomedical Nested Named Entity Linking at BioASQ 2025}. In: Faggioli, G., Ferro, N., Rosso, P., Spina, D. (eds.) CLEF 2025 Working Notes (2025)

\bibitem{Lin2004}
Lin, C.Y.: {{ROUGE}: A package for automatic evaluation of summaries}. In: {Proceedings of the ACL workshop `Text Summarization Branches Out'}. pp. 74--81. Barcelona, Spain (2004)

\bibitem{sapbert}
Liu, F., Shareghi, E., Meng, Z., Basaldella, M., Collier, N.: Self-alignment pretraining for biomedical entity representations. In: Proceedings of the 2021 Conference of the North American Chapter of the Association for Computational Linguistics: Human Language Technologies. pp. 4228--4238. ACL, Online (Jun 2021). \doi{10.18653/v1/2021.naacl-main.334}

\bibitem{sapbert-multilingual}
Liu, F., Vuli{\'c}, I., Korhonen, A., Collier, N.: Learning domain-specialised representations for cross-lingual biomedical entity linking. In: Proceedings of the 59th Annual Meeting of the Association for Computational Linguistics and the 11th International Joint Conference on Natural Language Processing (Volume 2: Short Papers). pp. 565--574. ACL, Online (Aug 2021). \doi{10.18653/v1/2021.acl-short.72}, \url{https://aclanthology.org/2021.acl-short.72/}

\bibitem{bionnel-LYX-DMIIP-FDU}
Liu, Y.: {LYX\_DMIIP\_FDU at BioASQ 2025: Utilizing BERT embeddings for biomedical text mining}. In: Faggioli, G., Ferro, N., Rosso, P., Spina, D. (eds.) CLEF 2025 Working Notes (2025)

\bibitem{NERELBIO}
Loukachevitch, N., Manandhar, S., Baral, E., Rozhkov, I., Braslavski, P., Ivanov, V., Batura, T., Tutubalina, E.: {NEREL-BIO: A Dataset of Biomedical Abstracts Annotated with Nested Named Entities}. Bioinformatics  (04 2023). \doi{10.1093/bioinformatics/btad161}, btad161

\bibitem{NEREL-BIO-COLING-2024}
Loukachevitch, N., Sakhovskiy, A., Tutubalina, E.: Biomedical concept normalization over nested entities with partial {UMLS} terminology in {R}ussian. In: Proceedings of the 2024 Joint International Conference on Computational Linguistics, Language Resources and Evaluation (LREC-COLING 2024). pp. 2383--2389. ELRA and ICCL, Torino, Italia (May 2024), \url{https://aclanthology.org/2024.lrec-main.213/}

\bibitem{malakasiotis2020evaluation}
Malakasiotis, P., Pavlopoulos, I., Androutsopoulos, I., Nentidis, A.: Evaluation measures for task b. Tech. rep., Tech. rep. BioASQ (2022), \url{http://participants-area.bioasq.org/Tasks/b/eval\_meas\_2022}

\bibitem{BioASQ2025taskGutBrainIE}
Martinelli, M., Silvello, G., Bonato, V., Di~Nunzio, G.M., Ferro, N., Irrera, O., Marchesin, S., Menotti, L., Vezzani, F.: {Overview of GutBrainIE@CLEF 2025: Gut-Brain Interplay Information Extraction}. In: Faggioli, G., Ferro, N., Rosso, P., Spina, D. (eds.) CLEF 2025 Working Notes (2025)

\bibitem{team_DS-GT-bioasq-task6_paper_gbie}
Mehta, R.: {Enhancing Biomedical Named Entity Recognition using GLiNER-BioMed with Targeted Dictionary-Based Post-processing for BioASQ 2025 task 6}. In: Faggioli, G., Ferro, N., Rosso, P., Spina, D. (eds.) CLEF 2025 Working Notes (2025)

\bibitem{BioASQECIR2025}
Nentidis, A., Katsimpras, G., Krithara, A., Krallinger, M., Ortega, M.R., Loukachevitch, N., Sakhovskiy, A., Tutubalina, E., Tsoumakas, G., Giannakoulas, G., Bekiaridou, A., Samaras, A., Di~Nunzio, G.M., Ferro, N., Marchesin, S., Menotti, L., Silvello, G., Paliouras, G.: {BioASQ at CLEF2025: The Thirteenth Edition of the Large-Scale Biomedical Semantic Indexing and Question Answering Challenge}. In: Advances in Information Retrieval. pp. 407--415. Springer Nature Switzerland, Cham (2025)

\bibitem{nentidis2023results}
Nentidis, A., Katsimpras, G., Krithara, A., Lima~L{\'o}pez, S., Farr{\'e}-Maduell, E., Gasco, L., Krallinger, M., Paliouras, G.: {Overview of BioASQ 2023: The Eleventh BioASQ Challenge on Large-Scale Biomedical Semantic Indexing and Question Answering}. In: Arampatzis, A., Kanoulas, E., Tsikrika, T., Vrochidis, S., Giachanou, A., Li, D., Aliannejadi, M., Vlachos, M., Faggioli, G., Ferro, N. (eds.) Experimental IR Meets Multilinguality, Multimodality, and Interaction. pp. 227--250. Springer Nature Switzerland, Cham (2023)

\bibitem{BioASQ2024overview}
Nentidis, A., Katsimpras, G., Krithara, A., Lima-López, S., Farré-Maduell, E., Krallinger, M., Loukachevitch, N., Davydova, V., Tutubalina, E., Paliouras, G.: {Overview of BioASQ 2024: The twelfth BioASQ challenge on Large-Scale Biomedical Semantic Indexing and Question Answering}. In: Experimental IR Meets Multilinguality, Multimodality, and Interaction. Proceedings of the Fifteenth International Conference of the CLEF Association (CLEF 2024) (2024)

\bibitem{nentidis2023ceur}
Nentidis, A., Katsimpras, G., Krithara, A., Paliouras, G.: {Overview of BioASQ Tasks 11b and Synergy11 in CLEF2023}. In: CEUR Workshop Proceedings (2023)

\bibitem{BioASQ2024task12bSynergy}
Nentidis, A., Katsimpras, G., Krithara, A., Paliouras, G.: {Overview of BioASQ Tasks 12b and Synergy12 in CLEF2024}. In: Faggioli, G., Ferro, N., Galuščáková, P., García Seco~de Herrera, A. (eds.) CLEF Working Notes (2024)

\bibitem{BioASQ2025task13bSynergy}
Nentidis, A., Katsimpras, G., Krithara, A., Paliouras, G.: {Overview of BioASQ Tasks 13b and Synergy13 in CLEF2025}. In: Faggioli, G., Ferro, N., Rosso, P., Spina, D. (eds.) CLEF 2025 Working Notes (2025)

\bibitem{nentidis2021overview}
Nentidis, A., Katsimpras, G., Vandorou, E., Krithara, A., Gasco, L., Krallinger, M., Paliouras, G.: {Overview of BioASQ 2021: The Ninth BioASQ Challenge on Large-Scale Biomedical Semantic Indexing and Question Answering}. In: International Conference of the Cross-Language Evaluation Forum for European Languages. pp. 239--263. Springer (2021)

\bibitem{nentidis2022overview}
Nentidis, A., Katsimpras, G., Vandorou, E., Krithara, A., Miranda-Escalada, A., Gasco, L., Krallinger, M., Paliouras, G.: {Overview of BioASQ 2022: The Tenth BioASQ Challenge on Large-Scale Biomedical Semantic Indexing and Question Answering}. In: {Experimental IR Meets Multilinguality, Multimodality, and Interaction}. Springer (2022). \doi{10.1007/978-3-031-13643-6\_22}

\bibitem{nentidis2021ceur}
Nentidis, A., Katsimpras, G., Vandorou, E., Krithara, A., Paliouras, G.: {Overview of BioASQ Tasks 9a, 9b and Synergy in CLEF2021}. In: Proceedings of the 9th BioASQ Workshop A challenge on large-scale biomedical semantic indexing and question answering. CEUR Workshop Proceedings (2021), \url{http://ceur-ws.org/Vol-2936/paper-10.pdf}

\bibitem{nentidis2022ceur}
Nentidis, A., Katsimpras, G., Vandorou, E., Krithara, A., Paliouras, G.: {Overview of BioASQ Tasks 10a, 10b and Synergy10 in CLEF2022}. In: CEUR Workshop Proceedings. vol.~3180, pp. 171--178 (2022)

\bibitem{nentidis2024bioasq}
Nentidis, A., Krithara, A., Paliouras, G., Krallinger, M., Sanchez, L.G., Lima, S., Farre, E., Loukachevitch, N., Davydova, V., Tutubalina, E.: {BioASQ at CLEF2024: The Twelfth Edition of the Large-Scale Biomedical Semantic Indexing and Question Answering Challenge}. In: ECIR2024. pp. 490--497. Springer (2024)

\bibitem{Oord2018RepresentationLW}
van~den Oord, A., Li, Y., Vinyals, O.: Representation learning with contrastive predictive coding. ArXiv  \textbf{abs/1807.03748} (2018), \url{https://api.semanticscholar.org/CorpusID:49670925}

\bibitem{team_ataupd2425-pam_paper_gbie}
Pamio, L., Di~Nunzio, G.M.: {BioASQ task GutBrainIE 2025 Task 6: Comparing CRF vs BERT Models for Named Entity Recognition and Relation Extraction}. In: Faggioli, G., Ferro, N., Rosso, P., Spina, D. (eds.) CLEF 2025 Working Notes (2025)

\bibitem{panou_2025}
Panou, D., Dimopoulos, A., Koubarakis, M., Reczko, M.: {Harnessing Collective Intelligence of LLMs for Robust Biomedical QA: A Multi-Model Approach}. In: Faggioli, G., Ferro, N., Rosso, P., Spina, D. (eds.) CLEF 2025 Working Notes (2025)

\bibitem{bionnel-VerbaNex-AI-Lab}
Peña~Gnecco, D., Serrano, J., Puertas, E., Mart{\'\i}nez-Santos, J.C.: {Hybrid Re-ranking for Biomedical Entity Linking using SapBERT Embeddings: A High-Performance System for BioNNE-L 2025-1}. In: Faggioli, G., Ferro, N., Rosso, P., Spina, D. (eds.) CLEF 2025 Working Notes (2025)

\bibitem{team_ataupd2425-gainer_paper_gbie}
Piron, S., Di~Nunzio, G.M.: {Named Entity Recognition with GLiNER and Relation Extraction with LLMs}. In: Faggioli, G., Ferro, N., Rosso, P., Spina, D. (eds.) CLEF 2025 Working Notes (2025)

\bibitem{BioASQ2025MultiClinSum}
Rodríguez-Ortega, M., Rodríguez-Lopez, E., Lima-López, S., Escolano, C., Melero, M., Pratesi, L., Vigil-Gimenez, L., Fernandez, L., Farré-Maduell, E., Krallinger, M.: {Overview of MultiClinSum task at BioASQ 2025: evaluation of clinical case summarization strategies for multiple languages: data, evaluation, resources and results.} In: Faggioli, G., Ferro, N., Rosso, P., Spina, D. (eds.) CLEF 2025 Working Notes (2025)

\bibitem{bionnel-overview-2025}
Sakhovskiy, A., Loukachevitch, N., Tutubalina, E.: {Overview of the BioASQ BioNNE-L Task on Biomedical Nested Entity Linking in CLEF 2025}. In: Faggioli, G., Ferro, N., Rosso, P., Spina, D. (eds.) {CLEF 2025 Working Notes} (2025)

\bibitem{GEBERT-CLEF-2023}
Sakhovskiy, A., Semenova, N., Kadurin, A., Tutubalina, E.: Graph-enriched biomedical entity representation transformer. In: Experimental IR Meets Multilinguality, Multimodality, and Interaction. pp. 109--120. Springer Nature Switzerland, Cham (2023)

\bibitem{BERGAMOT}
Sakhovskiy, A., Semenova, N., Kadurin, A., Tutubalina, E.: Biomedical entity representation with graph-augmented multi-objective transformer. In: Findings of the Association for Computational Linguistics: NAACL 2024. pp. 4626--4643. ACL, Mexico City, Mexico (Jun 2024). \doi{10.18653/v1/2024.findings-naacl.288}

\bibitem{Salton1990}
Salton, G., Buckley, C.: {Improving retrieval performance by relevance feedback}. Journal of the American Society for Information Science  \textbf{41}(4),  288--297 (jun 1990). \doi{10.1002/(SICI)1097-4571(199006)41:4<288::AID-ASI8>3.0.CO;2-H}

\bibitem{elisaTerumi_team}
Schneider, E.T.R., Schneider, F.H., Paraiso, E.C., Britto~Jr, A.S., Cruz, R.M.O.: {MedGemma-Sum-Pt: A Lightweight Model for Portuguese Clinical Summarization}. In: Faggioli, G., Ferro, N., Rosso, P., Spina, D. (eds.) CLEF 2025 Working Notes (2025)

\bibitem{Stachura25}
Stachura, D., Konieczna, J., Nowak, A.: {Are Smaller Open-Weight LLMs Closing the Gap to Proprietary Models for Biomedical Question Answering? }. In: Faggioli, G., Ferro, N., Rosso, P., Spina, D. (eds.) CLEF 2025 Working Notes (2025)

\bibitem{biosyn}
Sung, M., Jeon, H., Lee, J., Kang, J.: Biomedical entity representations with synonym marginalization. In: Proceedings of the 58th Annual Meeting of the Association for Computational Linguistics. pp. 3641--3650. ACL, Online (Jul 2020). \doi{10.18653/v1/2020.acl-main.335}

\bibitem{Tang25}
Tang, J., Yang, H., Xiong, K., Li, H., Quaresma, P., Yu, H., Zhang, W., Song, M., Jiang, Y.: {Applying DeepSeek to BioASQ Task 13B: Using Supervised Fine-Tuning and Few-Shot Learning }. In: Faggioli, G., Ferro, N., Rosso, P., Spina, D. (eds.) CLEF 2025 Working Notes (2025)

\bibitem{team_NLPatVCU_paper_gbie}
Taylor, S., Dil, C., Shah, A., Jannat, Oldham, C., Upadhyay, A., Varughese, J., Yazbeck, N., McInnes, B.T.: {NLP@VCU at BioASQ2025: Information Extraction on the GutBrainIE dataset}. In: Faggioli, G., Ferro, N., Rosso, P., Spina, D. (eds.) CLEF 2025 Working Notes (2025)

\bibitem{tedeschi-etal-2021-wikineural-multilingual-ner}
Tedeschi, S., Maiorca, V., Campolungo, N., Cecconi, F., Navigli, R.: {W}iki{NE}u{R}al: {C}ombined neural and knowledge-based silver data creation for multilingual {NER}. In: Findings of the Association for Computational Linguistics: EMNLP 2021. pp. 2521--2533. ACL, Punta Cana, Dominican Republic (Nov 2021). \doi{10.18653/v1/2021.findings-emnlp.215}

\bibitem{Tsatsaronis2015}
Tsatsaronis, G., Balikas, G., Malakasiotis, P., Partalas, I., Zschunke, M., Alvers, M.R., Weissenborn, D., Krithara, A., Petridis, S., Polychronopoulos, D., Almirantis, Y., Pavlopoulos, J., Baskiotis, N., Gallinari, P., Artieres, T., Ngonga, A., Heino, N., Gaussier, E., Barrio-Alvers, L., Schroeder, M., Androutsopoulos, I., Paliouras, G.: {An overview of the BIOASQ large-scale biomedical semantic indexing and question answering competition}. BMC Bioinformatics  \textbf{16}, ~138 (2015)

\bibitem{multi-embedding-prompt-driven}
Vachharajani, P.: Multilingual embedding and prompt-driven approaches for named entity recognition, entity linking, and clinical code prediction in greek discharge summaries. In: Faggioli, G., Ferro, N., Rosso, P., Spina, D. (eds.) CLEF 2025 Working Notes (2025)

\bibitem{pjmathematician_team}
Vachharajani, P.: {pjmathematician at MultiClinSUM 2025: A Novel Automated Prompt Optimization Framework for Multilingual Clinical Summarization.} In: Faggioli, G., Ferro, N., Rosso, P., Spina, D. (eds.) CLEF 2025 Working Notes (2025)

\bibitem{enigma-elcardiocc}
Velichkov, B., Datseris, A., Vassileva, S., Boytcheva, S.: {Enigma @ ElCardioCC: Bridging NER and ICD-10 Entity Linking - A Hybrid Method for Greek Clinical Narratives}. In: Faggioli, G., Ferro, N., Rosso, P., Spina, D. (eds.) CLEF 2025 Working Notes (2025)

\bibitem{Verma25}
Verma, S., Jiang, F., Xue, X.: {Beyond Retrieval: Ensembling Cross-Encoders and GPT Rerankers with LLMs for Biomedical QA }. In: Faggioli, G., Ferro, N., Rosso, P., Spina, D. (eds.) CLEF 2025 Working Notes (2025)

\bibitem{yang2016learning}
Yang, Z., Zhou, Y., Nyberg, E.: Learning to answer biomedical questions: {OAQA} at {B}io{ASQ} 4{B}. In: Kakadiaris, I.A., Paliouras, G., Krithara, A. (eds.) Proceedings of the Fourth {B}io{ASQ} workshop. pp. 23--37. ACL, Berlin, Germany (Aug 2016). \doi{10.18653/v1/W16-3104}

\bibitem{linkbert}
Yasunaga, M., Leskovec, J., Liang, P.: {LinkBERT: Pretraining Language Models with Document Links}. In: Association for Computational Linguistics (ACL) (2022)

\bibitem{zaratiana-etal-2024-gliner}
Zaratiana, U., Tomeh, N., Holat, P., Charnois, T.: {GL}i{NER}: Generalist model for named entity recognition using bidirectional transformer. In: Duh, K., Gomez, H., Bethard, S. (eds.) Proceedings of the 2024 Conference of the North American Chapter of the Association for Computational Linguistics: Human Language Technologies (Volume 1: Long Papers). pp. 5364--5376. ACL, Mexico City, Mexico (Jun 2024). \doi{10.18653/v1/2024.naacl-long.300}

\bibitem{zhang2020bertscore}
Zhang, T., Kishore, V., Wu, F., Weinberger, K.Q., Artzi, Y.: {BERTScore: Evaluating Text Generation with BERT}. In: International Conference on Learning Representations (ICLR) (2020), \url{https://arxiv.org/abs/1904.09675}

\bibitem{zhou2021atlop}
Zhou, W., Huang, K., Ma, T., Huang, J.: Document-level relation extraction with adaptive thresholding and localized context pooling. In: Proceedings of the AAAI Conference on Artificial Intelligence (2021)

\bibitem{roberta}
Zhuang, L., Wayne, L., Ya, S., Jun, Z.: A robustly optimized {BERT} pre-training approach with post-training. In: Li, S., Sun, M., Liu, Y., Wu, H., Liu, K., Che, W., He, S., Rao, G. (eds.) Proceedings of the 20th Chinese National Conference on Computational Linguistics. pp. 1218--1227. Chinese Information Processing Society of China, Huhhot, China (Aug 2021), \url{https://aclanthology.org/2021.ccl-1.108/}

\end{thebibliography}

\end{document}